\pdfoutput=1
\documentclass[10pt,twocolumn,letterpaper]{article}

\usepackage{iccv}
\usepackage{times}
\usepackage{algorithm}
\usepackage{algorithmic}
\usepackage{epsfig}
\usepackage{graphicx}
\usepackage{epstopdf}
\usepackage{amsmath}
\usepackage{amssymb}
\usepackage{multirow}
\usepackage{array}

\usepackage[pagebackref=true,breaklinks=true,letterpaper=true,colorlinks=false,bookmarks=false,backref=false]{hyperref}

\iccvfinalcopy % *** Uncomment this line for the final submission

\begin{document}

%%%%%%%%% TITLE
\title{LooseCut: Interactive Image Segmentation with Loosely Bounded Boxes}

\author{
\begin{tabular}{>{\centering}p{0.99\textwidth}}
Hongkai Yu\textsuperscript{\dag{}},Youjie Zhou\textsuperscript{\dag{}}, 
Hui Qian\textsuperscript{\ddag{}}, Min Xian\textsuperscript{*}, 
Yuewei Lin\textsuperscript{\dag{}}, Dazhou Guo\textsuperscript{\dag{}}, Kang Zheng\textsuperscript{\dag{}}, \tabularnewline
Kareem  Abdelfatah\textsuperscript{\dag{}} and Song Wang\textsuperscript{\dag{}}\tabularnewline
\tabularnewline
\textsuperscript{\dag{}}Department of Computer Science \& Engineering,University of South Carolina, Columbia, SC 29208\tabularnewline
\textsuperscript{\ddag{}}College of Computer Science, Zhejiang University, Hangzhou, China\tabularnewline
\textsuperscript{*}Department of Computer Science, Utah State University, Logan, UT 84341\tabularnewline
\texttt{\small{} \{yu55, zhou42\}@email.sc.edu, qianhui@zju.edu.cn, min.xian@aggiemail.usu.edu,}\tabularnewline
\texttt{\small{} \{lin59, guo22, zheng37\}@email.sc.edu, ker00@fayoum.edu.eg and songwang@cec.sc.edu} \tabularnewline
\end{tabular}
}
\maketitle

%%%%%%%%% ABSTRACT
\begin{abstract}
    One popular approach to interactively segment the foreground object of
  interest from an image is to annotate a bounding box that covers the
  foreground object. Then, a binary labeling is performed to achieve a refined segmentation.
  One major issue of the existing algorithms for such interactive image segmentation is their preference of an input bounding
  box that tightly encloses the foreground object. This increases the annotation
  burden, and prevents these algorithms from utilizing automatically detected
  bounding boxes. In this paper, we develop a new LooseCut algorithm that can
  handle cases where the input bounding box only loosely covers the foreground
  object. We propose a new Markov Random Fields (MRF) model for segmentation with loosely bounded boxes, including a global
  similarity constraint to better distinguish the foreground
  and background, and an additional energy term to encourage consistent labeling of similar-appearance pixels.
  This MRF model is then solved by an iterated max-flow algorithm. In the experiments, we evaluate LooseCut in three
  publicly-available image datasets, and compare its performance against several
  state-of-the-art interactive image segmentation algorithms. We also show that
  LooseCut can be used for enhancing the performance of unsupervised video
  segmentation and image saliency detection.
\end{abstract}

%%%%%%%%% BODY TEXT
\section{Introduction}
Accurately segmenting a foreground object of interest from an image with
convenient human interactions plays a central role in image and video
editing. One widely used interaction is to annotate a bounding box around the
foreground object. On one hand, this input bounding box provides the spatial
location of the foreground. On the other hand, based on the image information
within and outside this bounding box, we can have an initial estimation of the
appearance models of the foreground and background, with which a binary labeling
is finally performed to achieve a refined segmentation of the foreground and
background ~\cite{rother2004grabcut, tang2013onecut, tang2014pseudo,
  wu2014milcut, lempitsky2009image, kass1988snakes}.

% to-do: SlackCut ->LooseCut in figure. Delete the bottom example when there is no space. Change main text too.
\begin{figure}[thp]
  \includegraphics[width=1\columnwidth]{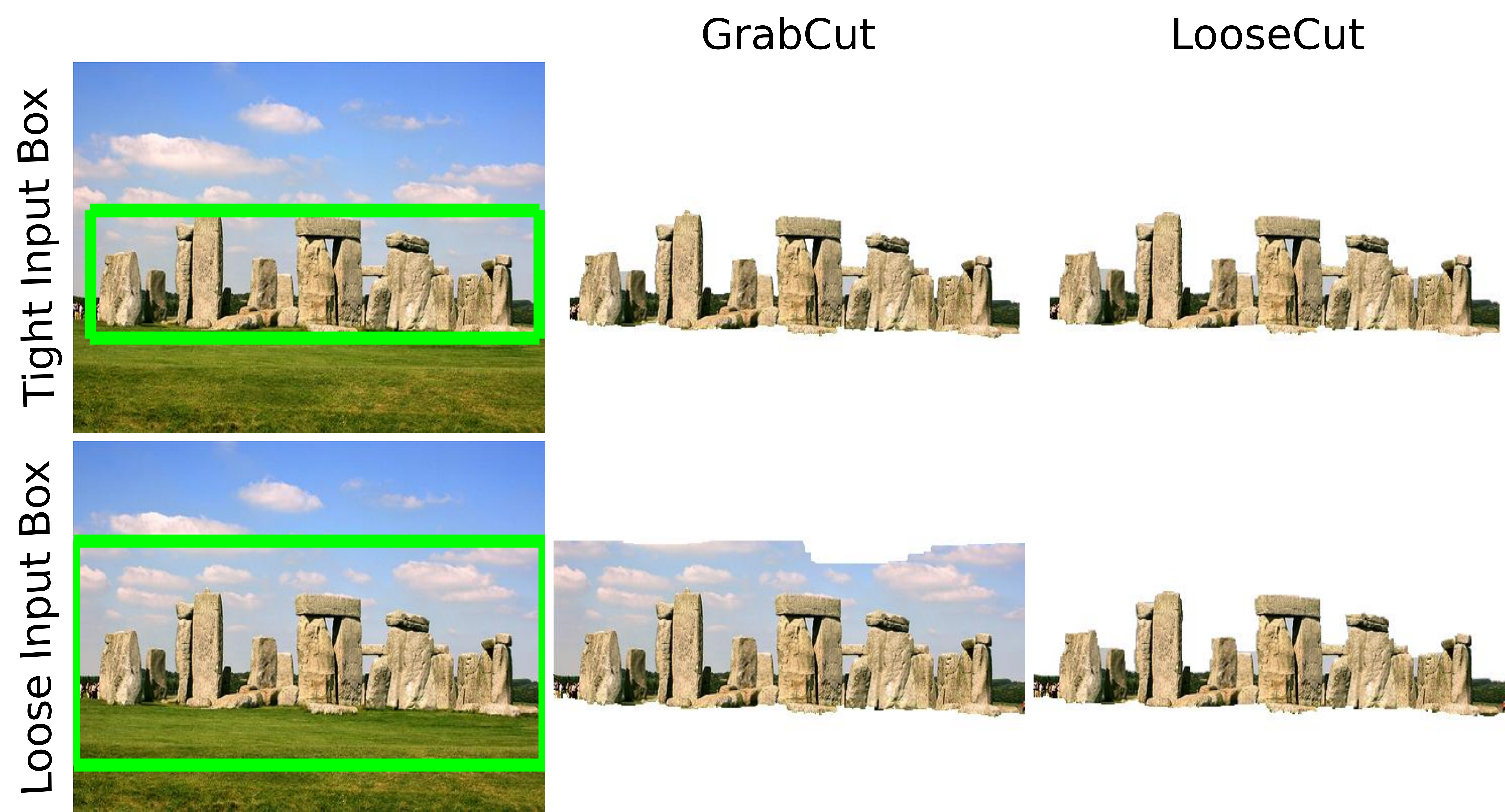}
  \caption{Sample results from GrabCut and the proposed LooseCut with tightly
    and loosely bounded boxes.}
\vspace{-1 em}
  \label{fig1}
\end{figure}

However, due to the complexity of the object boundary and appearance, most of the
existing methods of this kind prefer the input bounding box to tightly enclose
the foreground object. An example is shown in Fig.~\ref{fig1}, where the widely
used GrabCut \cite{rother2004grabcut} algorithm fails when the bounding box does
not tightly cover the foreground object. The preference of a tight bounding box
increases the burden of the human interaction, and moreover it prevents these
algorithms from utilizing automatically generated bounding boxes, such as boxes
from object proposals \cite{alexe2010object, zitnick2014edge, zhou2015feature},
that are usually not guaranteed to tightly cover the foreground object. In this
paper, we focus on developing a new LooseCut algorithm that can accurately
segment the foreground object with loosely-bounded boxes.

\begin{figure}[htbp]
\begin{center}
  \includegraphics[width=1\columnwidth]{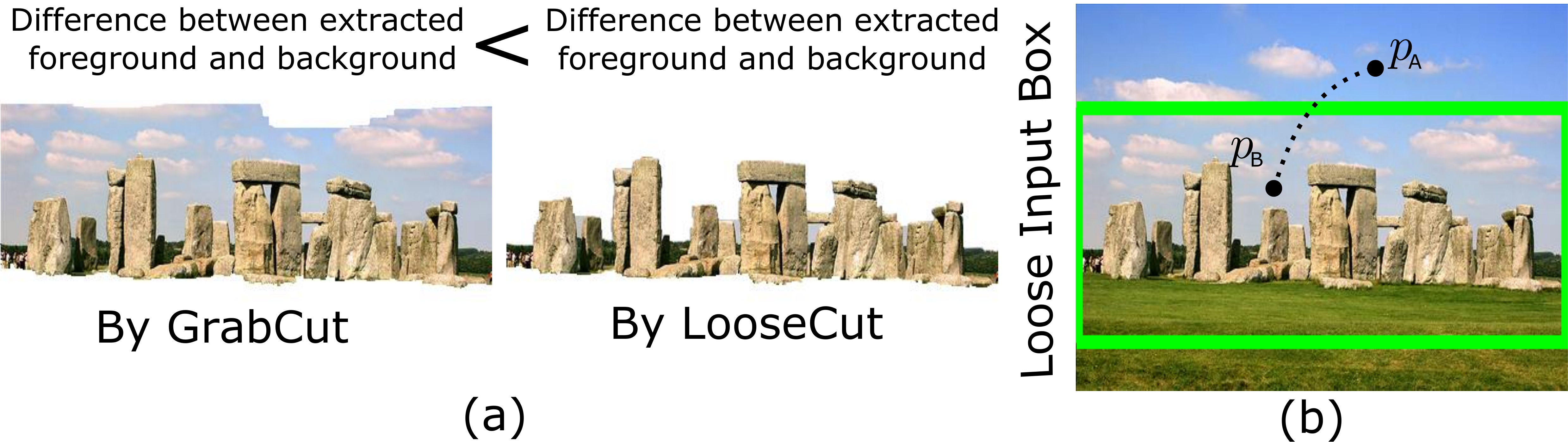}
\end{center}
\vspace{-1 em}
\caption{An illustration of the two strategies used in the proposed LooseCut algorithm.
(a) By emphasizing their appearance difference, foreground and background are better separated even with a loosely bounding box.
(b) By encouraging label consistency of similar-appearance pixels,
background pixel $P_{B}$ inside the loosely bounded box is correctly labeled as background due to its
appearance similarity to the background pixel $P_{A}$ outside the bounding box.}
\label{fig:Motivation}
\vspace{-1 em}
\end{figure}

A loosely bounded box may contain more background than a tightly bounded box. As a result,
the initial appearance model of the foreground is highly inaccurate by using the pixels within
the bounding box. This may substantially reduce the segmentation performance
as shown by the Grabcut result in Fig.~\ref{fig1}. In this paper, we propose two strategies
to address this problem. First, we explicitly emphasize the appearance difference between the
foreground and background models. Second, we explicitly encourage the consistent labeling to the similar-appearance
pixels, either adjacent or non-adjacent. These two strategies can help identify the background pixels
within the bounding box, as shown in Fig.~\ref{fig:Motivation}.

In this paper, we follow GrabCut by formulating the foreground/background segmentation as a binary labeling over an
MRF built upon the image grid, and the appearances of the foreground and background are described by two Gaussian Mixture Models (GMMs).
More specifically, we add a \emph{global similarity constraint} and a \emph{label consistency term} to the MRF energy
to implement the above mentioned two strategies. Finally, we solve the proposed MRF model using
an iterated max-flow algorithm. In the experiments, we evaluate the proposed LooseCut in three publicly-available image datasets, and compare its performance against several
state-of-the-art interactive image segmentation algorithms. We also show that LooseCut can be used for enhancing the performance of unsupervised video
segmentation and image saliency detection.

% to-do: use \ref labels for the section number.
The remainder of the paper is organized as follows. Section \ref{sec:related}
reviews the related work. Section \ref{sec:algo} describes the proposed LooseCut algorithm in
detail. Section \ref{sec:exp} reports the experimental results, followed by a
briefly conclusion in Section \ref{sec:end}.
% -------------------------------------------------------------------------------
\section{Related Work\label{sec:related}}
In recent years, interactive image segmentation based on input bounding boxes
have drawn much attention in the computer vision and graphics community,
resulting in a number of effective algorithms~\cite{rother2004grabcut,
  tang2013onecut,tang2014pseudo,
  wu2014milcut,lempitsky2009image,kass1988snakes}. Starting from the classical
GrabCut algorithm, many of these algorithms use graph cut models: the input
image is modeled by a graph and the foreground/background segmentation is then
modeled by a binary graph cut that minimizes a pre-defined energy
function~\cite{boykov2001interactive}. In GrabCut~\cite{rother2004grabcut},
initial appearance models of the foreground and background are estimated using
the image information within and outside the bounding box. A binary MRF model is
then applied to label each pixel as the foreground or background, based on which
the appearance models of the foreground and background are re-estimated. This
process is repeated until convergence. As illustrated in Fig.~\ref{fig1}, the
performance of GrabCut is highly dependent on the initial estimation of the
appearance models of the foreground and background, which might be very poor
when the input bounding box does not tightly cover the foreground object. The
LooseCut algorithm developed in this paper also follows the general procedure
introduced in GrabCut, but introduce a new constraint and a new energy term
to the MRF model to specifically handle the loosely-bounded boxes.

PinPoint~\cite{lempitsky2009image} is another MRF-based algorithm for
interactive image segmentation with a bounding box. It incorporates a topology
prior derived from geometry properties of the bounding box and encourages the
segmented foreground to be tightly enclosed by the bounding box. Therefore, its
performance gets much worse with a loosely bounded box. Also using an MRF model,
OneCut~\cite{tang2013onecut} is recently developed for interactive image segmentation.
Its main contribution is to incorporate an MRF energy term that reflects
the appearance overlap between foreground and background histograms. As shown in the
latter experiments, the $L_1$-distance based appearance overlap used in
OneCut is still insufficient to handle loosely-bounded boxes. In~\cite{tang2014pseudo},
a pPBC algorithm is developed for
interactive image segmentation using an efficient parametric pseudo-bound
optimization strategy. However, in our experiment shown in Section \ref{sec:exp},
pPBC still cannot give satisfactory segmentation results when the input bounding box is loose.

Other than using the MRF model, MILCut~\cite{wu2014milcut} formulates the
interactive image segmentation as a multiple instance learning problem by
generating positive bags along the sweeping lines within the bounding
box. MILCut may not generate the desirable
positive bags along the sweeping lines for a loosely bounded box.
Active contour~\cite{kass1988snakes}
takes the input bounding box as an initial contour and iteratively deforms it
toward the boundary of the foreground object. Due to its sensitivity to image
noise, active contour usually requires the initial contour to be close to the
underlying foreground object boundary.

% to-do:  may put this paragraph into experiments
% However, loose bounding box always exists when these algorithms are applied in
% unsupervised image/video segmentation and saliency detection. For automatic
% salient region segmentation \cite{Cheng2015PAMI}, GrabCut is applied to the
% dilation of naive segmentation result of saliency map to improve the
% unsupervised segmentation. To keep a high recall, the threshold for naive
% segmentation is always quite low, which generates a relatively loose
% segmentation result covering the object as the bounding box for GrabCut. In
% unsupervised co-segmentation area, GrabCut is also applied to co-segment the
% common objects after automatically obtaining some loose bounding boxes around
% the common objects \cite{yu2014unsupervised} \cite{gao2013mutual}. Most
% saliency models focus on the local or global contrast, while \cite{li2013co}
% generates a saliency map from another perspective. Using predefined windows,
% GrabCut is applied to the windows and finally saliency map is constructed by
% multiscale segmentation voting. Some proposal algorithms like
% \cite{alexe2010object} \cite{zitnick2014edge} \cite{zhou2015feature} can be
% used to substitute the windows. The predefined windows and proposals are all
% loose bounding boxes compared to object size.
%-------------------------------------------------------------------------------
%----------------------------------------------------------------------------
\section{Proposed Approach\label{sec:algo}}
In this section, we first briefly review the classical GrabCut algorithm and then explain the proposed LooseCut algorithm.
	
\subsection{GrabCut\label{sec:grabcut}}
GrabCut~\cite{rother2004grabcut} actually performs a binary labeling to each
pixel using an MRF model. Let $X=\left\{ x_{i}\right\} _{i=1}^{n}$ be the binary
labels at each pixel $i$, where $x_{i}=1$ if $i$ is in foreground $x_{i}=0$ if
$i$ is in background and let $\theta=(M_f, M_b)$ denotes the appearance models including foreground GMM $M_f$ and background GMM $M_b$. Grabcut seeks an optimal labeling that minimizes
\begin{equation}
  E_{GC}\left(X, \theta \right)=\sum_{i}D\left(x_{i}, \theta \right)+\sum_{{i,j}\in\mathcal{N}}V\left(x_{i},x_{j}\right)\textrm{,}\label{eq:GC}
\end{equation}
\noindent where $\mathcal{N}$ defines a pixel neighboring system, e.g.,
4-neighbor or 8-neighbor connectivity. The unary term $D\left(x_{i}, \theta \right)$
measures the cost of labeling pixel $i$ as foreground or background based on the appearance models $\theta$.
The pairwise term $V\left(x_{i},x_{j}\right)$ enables the smoothness of the labels
by penalizing discontinuity among the neighboring pixels with different labels.  Max-flow algorithm
\cite{boykov2001interactive} is usually used for solving this MRF optimization
problem. GrabCut takes the following steps to achieve the binary image segmentation with an input bounding box:
\begin{enumerate}
\item Estimating initial appearance models $\theta$, using the pixels inside and outside the bounding
  box respectively.
\item Based on the current appearance models $\theta$, quantizing the foreground and background
  likelihood of each pixel and using it to define the unary term
  $D\left(x_{i}, \theta \right)$. And solve for the optimal labeling that minimizes Eq.~(\ref{eq:GC}).
\item Based on the obtained labeling $X$, refining $\theta$ and going back to Step 2. Repeating this process until convergence.
\end{enumerate}
%---------------------------------------------------------------------------------
\subsection{MRF Model for LooseCut}
Following the MRF model used in GrabCut, the proposed LooseCut takes the following MRF energy function:
%\begin{equation}
%  E\left(X\right)=E_{GC}\left(X\right)+ E_{G}\left(X\right)+\beta E_{H}\left(X\right)\textrm{,}\label{eq:Full}
%\end{equation}
\vspace{-1 em}
\begin{equation}
\begin{split}
    E(X, \theta)=E_{GC}(X, \theta)+ \beta E_{LC}(X) \textrm{,}
\end{split}
\label{eq:Full}	
\end{equation}
where $E_{GC}$ is the GrabCut energy given in Eq.~(\ref{eq:GC}), and $E_{LC}$ is an energy term for encouraging label consistency, weighted by $\beta>0$.
In minimizing Eq.~(\ref{eq:Full}), we enforce a global similarity constraint to better estimate $\theta$ and distinguish the foreground and background.
In the following, we elaborate on the global similarity constraint and the label consistency term $E_{LC}(X)$.

%
% Since $M_b$ is more reliable in bounding box based interactive segmentation (outside of bounding box: background), our global similarity constraint can help to obtain better $M_f$ with small similarity to $M_b$ when estimating $\theta$. This global similarity constraint helps to minimize the unary energy term when the foreground and background are highly overlapped with loose bounding box, which indirectly leads to small similarity of extracted foreground and background. The label consistency energy term $E_{LC}$ simulates a farther correspondence in MRF, which softly encourages pixels with similar appearance to be assigned the same label even if they are not neighboring to each other. This property of label consistency could
%make segmentation more robust to loose bounding box.

%---------------------------------------------------------------------------------------------
\subsection{Global Similarity Constraint}
In this section, we define the proposed global similarity constraint.
Let $M_f$ have $K_f$ Gaussian components $M_f^i$ with means $\mu_f^i$ %%%%and weights $\pi_f^i$
, $i=1, 2, \cdots, K_f$ and $M_b$ have $K_b$ Gaussian components $M_b^j$
with means $\mu_b^j$ %%% and weights $\pi_b^j$
, $j=1, 2, \cdots, K_b$. For each Gaussian component $M_{f}^{i}$ in the foreground GMM $M_{f}$, we first find its nearest Gaussian component $M_{b}^{j(i)}$ in $M_{b}$ as

\begin{equation}
  j(i)=\arg\min_{j \in \{1,...,K_{b}\}}\left|\mu_{f}^{i}-\mu_{b}^{j}\right|\textrm{.}
\end{equation}
With this, we can define the similarity between the Gaussian component
$M_{f}^{i}$ and the entire background GMM $M_{b}$ as
\vspace{-1 em}

\begin{equation}
  %S\left(M_{f}^{i},M_{b}\right)=\frac{1}{\pi_{f}^{i}\left|\mu_{f}^{i}-\mu_{b}^{j(i)}\right|}\textrm{,}
  S\left(M_{f}^{i},M_{b}\right)=\frac{1}{\left|\mu_{f}^{i}-\mu_{b}^{j(i)}\right|}\textrm{,}
  \label{eq:sim}
\end{equation}
which is the inverse of the mean difference between $M_{f}^{i}$ and
its nearest Gaussian component in the background GMM. Then, we define
the global similarity function $Sim$ as
\begin{equation}	
Sim(M_{f},M_{b})=\sum_{i=1}^{K_{f}}S\left(M_{f}^{i},M_{b}\right).
\label{eq:global}
\end{equation}

Similar definition for GMM distance could be found in \cite{yu2014icip}. In the MRF minimization, we will enforce the global similarity $Sim(M_{f},M_{b})$ to be low (smaller than a threshold) in the step of estimating $\theta$ and details will be discussed in Section~\ref{sec:optimization}.
%---------------------------------------------------------------------------------
\subsection{Label Consistency Term $E_{LC}$}
%The pixels outside the bounding box are enforced as background labels in
%interactive segmentation.  We expect the inside-box pixels holding similar
%feature with outside-box pixels to be set the same label, namely background. We
%introduce a high order energy term to keep label consistency within foreground
%and background.
%
%High order energy defined on sets of pixels could overcome the disadvantage of
%sensitive pairwise smooth energy term and simulate the correspondence of
%long-distance pixels \cite{kohli2009robust}. Our idea is that the pixels
%belonging to the same cluster should be encouraged to be set the same label. The
%purpose that we use cluster as a clique to enforce label consistency is that we
%could include spatial and appearance information into cluster, which is more
%robust for label consistency.
%
%High order energy in MRF or Conditional Random Fields (CRF) obtains promising
%results in segmentation recently. Kohli \textit{et al.} \cite{kohli2009robust}
%proposes a high order pseudo-boolean potential to enforce label consistency on a
%clique/segment. Tang \textit{et al.} \cite{tang2013onecut} proposes a similar
%high order energy to penalize the overlap between the object and background in
%histogram bins. Jain \textit{et al.} \cite{jain2014supervoxel} introduces a high
%order supervoxel label consistency potential in MRF to improve object
%segmentation propagation in video.
%
%The high order energy term in proposed LooseCut algorithm is defined as

% to-do: put the review of high-order term to the related work.

\begin{figure}[htbp]
\begin{center}
  \includegraphics[width=0.6\columnwidth]{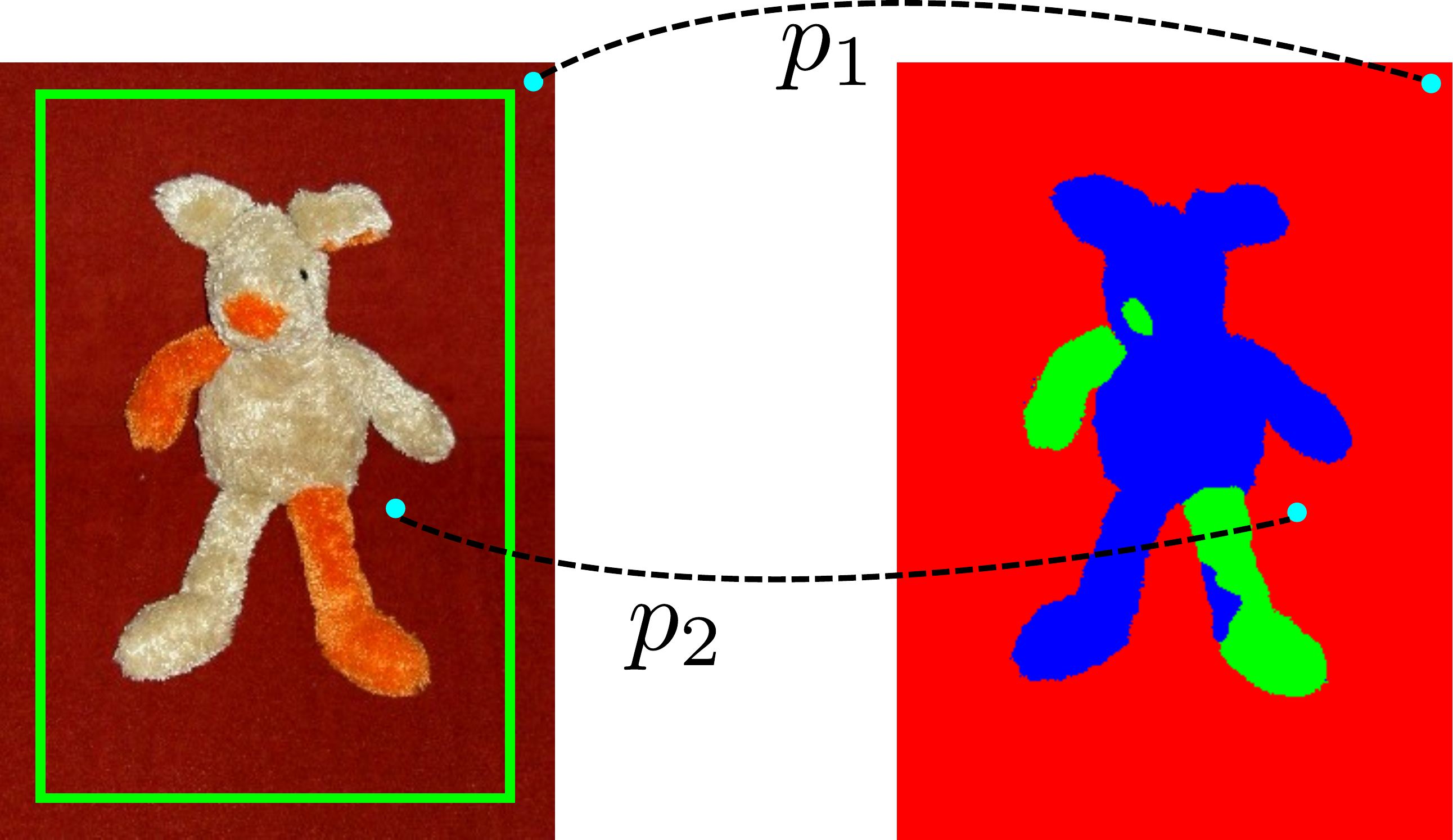}
\end{center}
\vspace{-1 em}
\caption{An illustration of the superpixel based clusters and label consistency. Three clusters are shown in different colors.}
\label{fig:high-order}
\vspace{-1 em}
\end{figure}

To encourage the label consistency of the similar-appearance pixels, either adjacent or
non-adjacent, we first cluster all the image pixels using a recent superpixel
algorithm~\cite{zhou2015multiscale} that preserves both feature and spatial consistency.
Following a \emph{K}-means-style procedure, this cluster algorithm partitions the image into a
set of compact superpixels and each resulting cluster is made up of one or more
superpixels. An example is shown in Fig.~\ref{fig:high-order}, where the region
color indicates the clusters: superpixels with the same color constitute a
cluster.

Let $C_k$ indicates the cluster $k$, and pixels belonging to $C_k$ should be
encouraged to be given the same label, e.g., $p_1$ and $p_2$ in Fig.~\ref{fig:high-order}.
To accomplish this, we set a cluster label $x_{C_k}$ (taking values 0 or 1)
for each cluster $C_k$ and define the label-consistency energy term as
\vspace{-0.5 em}
\begin{equation}
  E_{LC}(\mathbf{X}) = \sum_k\sum_{i\in C_k} \phi(x_{i} \neq x_{C_k})\textrm{,}
\label{eqHighOrder}	
\end{equation}
where $\phi(\cdot)$ is an indicator function taking 1 or 0 for true or false argument. In the proposed algorithm,
we will solve for both the pixel labels and cluster labels simultaneously in the MRF optimization.
%-------------------------------------------------------------------------

\subsection{Optimization\label{sec:optimization}}

In this section, we propose an algorithm to find the optimal binary labeling
that minimizes the energy function defined in Eq.~(\ref{eq:Full}), subject to the global similarity constraint.
Specifically, in each iteration, we first fix the labeling $X$ and optimize over $\theta$ by enforcing the global
similarity constraint on $Sim(M_f,M_b)$. After that, we fix $\theta$ and find an optimal $X$ that minimizes $E(X, \theta)$.
These two steps of optimization is repeated alternately until convergence or a preset maximum number of iterations is reached. 
As an initialization, we use the input bounding box to define a binary labeling $X$ in iteration 0.
% Experimental results will show that our algorithm could converge to a reasonable local minimum of $E(X, \theta)$. 
In the following, we elaborate on these two optimization steps.

\textbf{Fixing $X$ and Optimizing over $\theta$:} With fixed binary labeling $X$, we
can estimate $\theta$ using a standard EM-based clustering algorithm: All the pixels
with label 1 are taken for computing the foreground GMM $M_f$ and all the pixels with label 0 
are used for computing the background GMM $M_b$. We intentionally select $K_f$ and $K_b$ such that
$K=K_f-K_b>0$ since some background components are mixed to the foreground for the initial
$X$ defined by a loosely bounded box. For the obtained $M_f$ and $M_b$, we examine
whether the global similarity constraint is satisfied, i.e, $Sim(M_f,M_b)\leq \delta$ or not.
If this constraint is satisfied, we take the resulting $\theta$ and continue to the next step of optimization.
If this constraint is not satisfied, we further refine $M_f$ using the following algorithm:
\begin{enumerate}
\item  Calculate the similarity $S(M_f^i, M_b)$ between each Gaussian component of $M_f$ and $M_b$, by
following Eq.~(\ref{eq:sim}) and identify the $K$ Gaussian components of $M_f$
with the largest similarity to $M_b$. 
\item Among these $K$ components, if any one,
say $M_f^i$, does not satisfy $S(M_f^i, M_b) \leq \delta$, we delete it from $M_f$. 
\item After all the deletions, we use the remaining Gaussian components to construct an updated $M_f$. 
\end{enumerate}
This algorithm will ensure the updated $M_f$ and $M_b$ satisfies the global similarity constraint. 

%Minimizing the high order energy term is not easy. \cite{kohli2009robust} tries
%to enforce label consistency on a clique/segment by using two auxiliary nodes on
%a clique for optimization. \cite{tang2013onecut} simplifies the problem and
%treats each color histogram bin as a clique, so uses single auxiliary node on a
%color histogram bin for optimization. In \cite{tang2013onecut}, pixels belonging
%to the same color histogram bin are connected to an auxiliary node in graph and
%then max-flow algorithm is applied to minimize high order energy.

\begin{figure}[htbp]
\begin{center}
   \includegraphics[width=0.85\columnwidth]{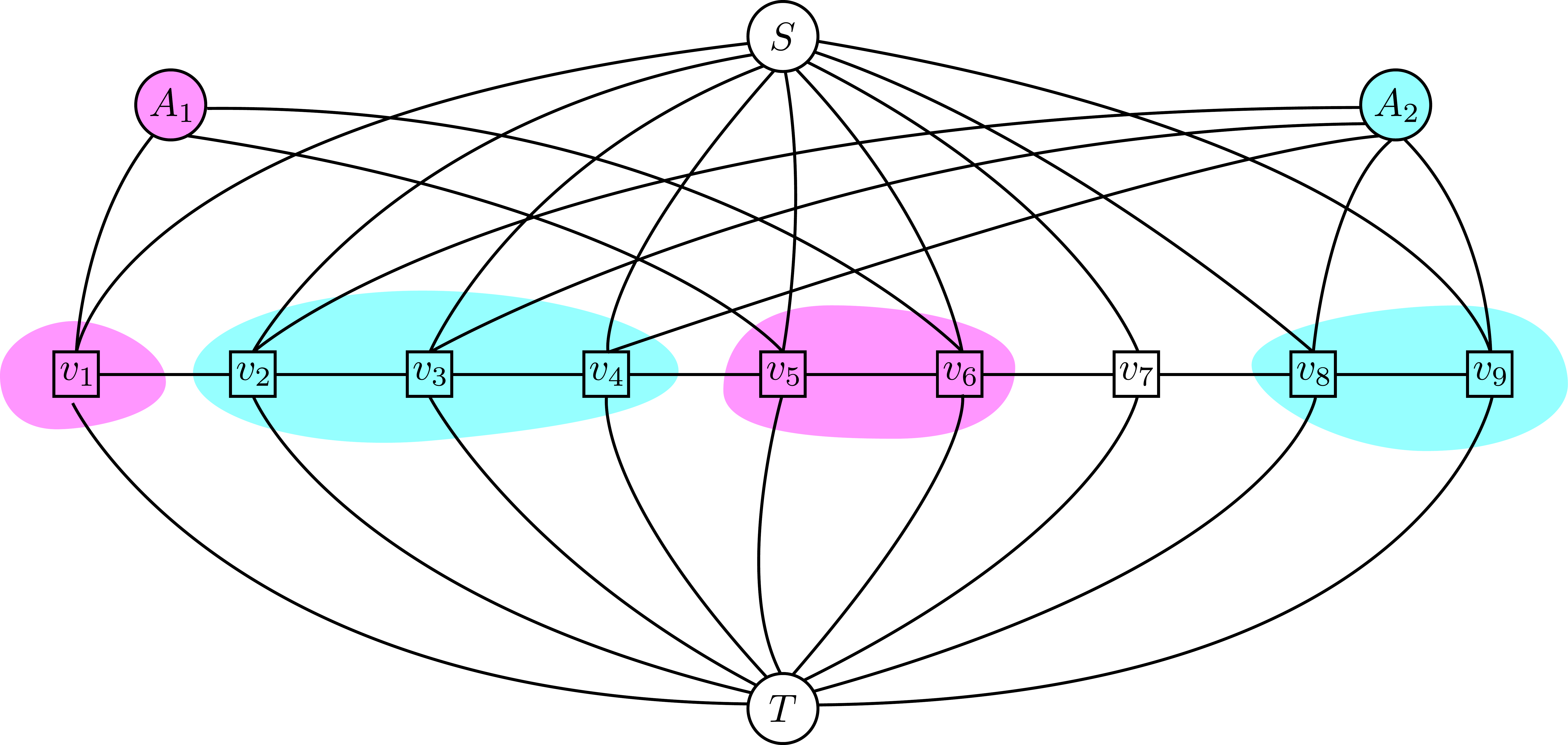}
\end{center}
\vspace{-1 em}
\caption{Graph construction for the step of optimizing over $X$ with a fixed $\theta$. $v_i$'s are the nodes
for pixels and $A_i$'s are the auxiliary nodes. $S$ and $T$ are the source and sink nodes. Same color nodes represent
a cluster.}
\label{fig2}
\vspace{-1 em}
\end{figure}

\textbf{Fixing $\theta$ and Optimizing over $X$}: Inspired by~\cite{kohli2009robust}
and~\cite{tang2013onecut}, we build an undirect graph with auxiliary nodes as shown in Fig.~\ref{fig2}
to find an optimal $X$ that minimizes the energy $E(X, \theta)$.
In this graph, each pixel is represented by a node. For each pixel cluster $C_k$, we construct an auxiliary node $A_k$ to represent it.
Edges are constructed to link the auxiliary node $A_k$ and the nodes that represent the pixels in $C_k$, with the
edge weight $\beta$ as used in Eq.~(\ref{eq:Full}). An example of the constructed graph is shown in Fig.~\ref{fig2}, 
where pink nodes $v_1$, $v_5$, and $v_6$ represent three pixels in a same cluster, which is represented by the 
auxiliary node $A_1$. All the nodes in blue represent another cluster. 
With a fixed $\theta$, we use the max-flow algorithm~\cite{boykov2001interactive} on this graph to seek an 
optimal $X$ that minimizes the energy $E(X, \theta)$.

% The cluster label $x_{C_k}$ is computed from the labels of all its adjacent pixels depends on the labels of $A_k$'s connected pixel nodes after cutting edges in optimization.

%
%For the edge $\beta$. Similar graph structure can be found in OneCut~\cite{tang2013onecut}, and we will show that our label-consistency term $E_{LC}$ can be modified to generate OneCut's energy term in Section ~\ref{sec:optimization}.

The graph constructed as in Fig.~\ref{fig2} is similar to the graph constructed in OneCut~\cite{tang2013onecut}. 
However, there are two major differences between the proposed algorithm and OneCut.
\begin{enumerate}
\item In OneCut, a color histogram is first constructed for the input image and then one auxiliary node is constructed 
for each histogram bin. All the pixels are then quantized into these bins and the pixels in each bin are then linked to its
corresponding auxiliary node. In this paper, we use superpixel-based clusters to define the auxiliary nodes.
\item The unary energy term in OneCut is different from the one in the proposed method and as a result,
we define the edge weights involving the source and sink nodes differently from OneCut. OneCut follows the ballooning 
technique: The weight is set to 1 for the edges between $S$ and any pixels inside the bounding box, and 0 otherwise; Similarly,
the weight is set to 0 for the edges between $T$ and any pixels in the bounding box, and $\infty$ otherwise. In the proposed algorithm,
the weights of the edges that are incident from $S$ or $T$ reflect the unary term in Eq.~(\ref{eq:Full}), which is
based on the appearance models $\theta$.
\end{enumerate}
With these two differences, OneCut seeks to minimize the $L_1$-distance based histogram overlap between the foreground and background.
This is different from the goal of the proposed algorithm: we seek better label consistency of the pixels in the same cluster
by using this graph structure. We will compare with OneCut in the latter experiments. The full LooseCut algorithm is summarized in Algorithm ~\ref{alg1}.

%If we use auxiliary nodes to represent histogram bins instead of superpixel based clusters, and set unary energy term to be ballooning (unary t-links inside box: 1 to $S$ and 0 to $T$, unary t-links outside box: 0 to $S$ and $\infty$ to $T$) and $\beta=1$ in Fig.~\ref{fig2}, and remove our global similarity constraint, minimizing $E_{LC}$ in Eq.~\ref{eqHighOrder} is actually the same as minimizing the $L_{1}$ overlap energy of foreground and background histograms (say, $E_{L_{1}}$) defined in OneCut. Suppose $k^{th}$ bin is split into foreground and background, generating $n^{f}_{k}$ and $n^{b}_{k}$ pixels respectively. OneCut prefers to cut min$ \{n^{f}_{k}, n^{b}_{k}\}$ number of links that connect the pixels in $k^{th}$ bin to the auxiliary node $A_{k}$, which exactly minimizes $E_{L_{1}}$ by repeating this in each bin, so more number of pixels in $k^{th}$ bin still connect to auxiliary node $A_{k}$, keeping same label with $x_{A_{k}}$. This is the same to minimize our label consistency energy $E_{LC}$. With different purpose, $E_{LC}$ in LooseCut encourages label consistency while $E_{L_{1}}$ in OneCut minimizes the overlap of foreground and background histograms. Even using the same graph structure, different settings in unary t-links and auxiliary nodes lead to different energy terms. Choosing superpixel based clusters instead of histogram bins is to consider feature and spatial consistency together.
 
\begin{algorithm}\caption{LooseCut}\label{alg1}
  \textbf{Input:} Image $I$, bounding box $B$, \# of clusters $N$
  \newline\textbf{Output:} \leftline{Binary labeling $X$ to pixels in $I$}
  \vspace{-1 em}
\begin{algorithmic}[1]
\STATE Construct $N$ superpixel based clusters using ~\cite{zhou2015multiscale}.
\STATE Create initial labeling $X$ using box $B$.
  \REPEAT
\STATE  Based on the current labeling $X$, estimate and update $\theta$ by enforcing $Sim(M_f, M_b)\leq\delta$.
\STATE  Construct the graph using the updated $\theta$ with $N$ auxiliary nodes as shown in Fig.~\ref{fig2}.
\STATE  Apply the max-flow algorithm~\cite{boykov2001interactive} to update labeling $X$ by minimizing $E(X, \theta)$.
  \UNTIL{Convergence or maximum iterations reached}
\end{algorithmic}
\label{alg1}
\end{algorithm}
\vspace{-1 em}

%-------------------------------------------------------------------------
\section{Experiments\label{sec:exp}}

To justify the proposed LooseCut algorithm, we conduct experiments on three
widely used image datasets -- the GrabCut dataset~\cite{rother2004grabcut}, the
Weizmann dataset~\cite{alpert2007image, borenstein2002class}, and the
iCoseg dataset~\cite{batra2010icoseg}, and compare its performance against
several state-of-the-art interactive image segmentation methods, including
GrabCut~\cite{rother2004grabcut}, OneCut~\cite{tang2013onecut},
MILCut~\cite{wu2014milcut}, and pPBC~\cite{tang2014pseudo}.  We also conduct
experiments to show the effectiveness of LooseCut in two applications:
unsupervised video segmentation and image saliency detection.

\textbf{Metrics}: As in~\cite{wu2014milcut} \cite{tang2013onecut}
\cite{lempitsky2009image}, we use \emph{Error Rate} to evaluate an interactive
image segmentation by counting the percentage of misclassified pixels inside the
bounding box. We also take the pixel-wise \emph{F-measure} as an evaluation
metric, which combines the precision and recall metrics in terms of the
ground-truth segmentation.

\textbf{Parameter Settings}: For the number of Gaussian components in GMMs,
$K_b$ is set to $5$ and $K_f$ is set to $6$. As discussed in Section \ref{sec:optimization},
$K=K_f-K_b=1$. To enforce the global similarity constraint, we delete $K=1$ component
in $M_f$. The number of clusters (auxiliary nodes in graph) is set to $N=16$. For the LooseCut
energy defined in Eq.~(\ref{eq:Full}), we consistently set $\beta=0.01$. The
unary term and binary term in Eq.~(\ref{eq:Full}) are the same as
in~\cite{rother2004grabcut} and RGB color features are used to construct the
GMMs. We set $\delta=0.02$ in deleting the foreground GMM component to enforce
the global similarity constraint. For all the comparison methods, we 
follow their default or recommended settings in their codes.

\begin{figure}[htbp]
\begin{center}
  \includegraphics[width=0.95\columnwidth]{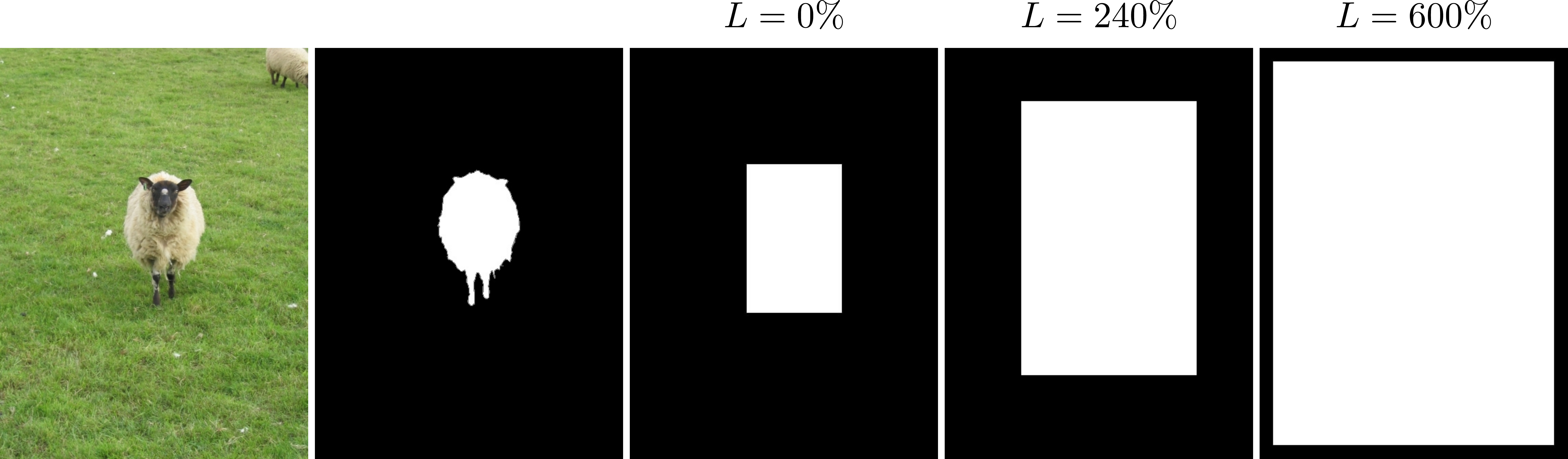}
\caption{Bounding boxes with different looseness. From left to right:
  image, ground-truth foreground, baseline bounding box and a series of
  bounding boxes with increased looseness. }
\label{figBox}
\end{center}
\end{figure}

% Youjie: Tune figure layout
\begin{figure*}[htbp]
\begin{centering}
\begin{tabular}{ccc}
\includegraphics[width=0.3\textwidth]{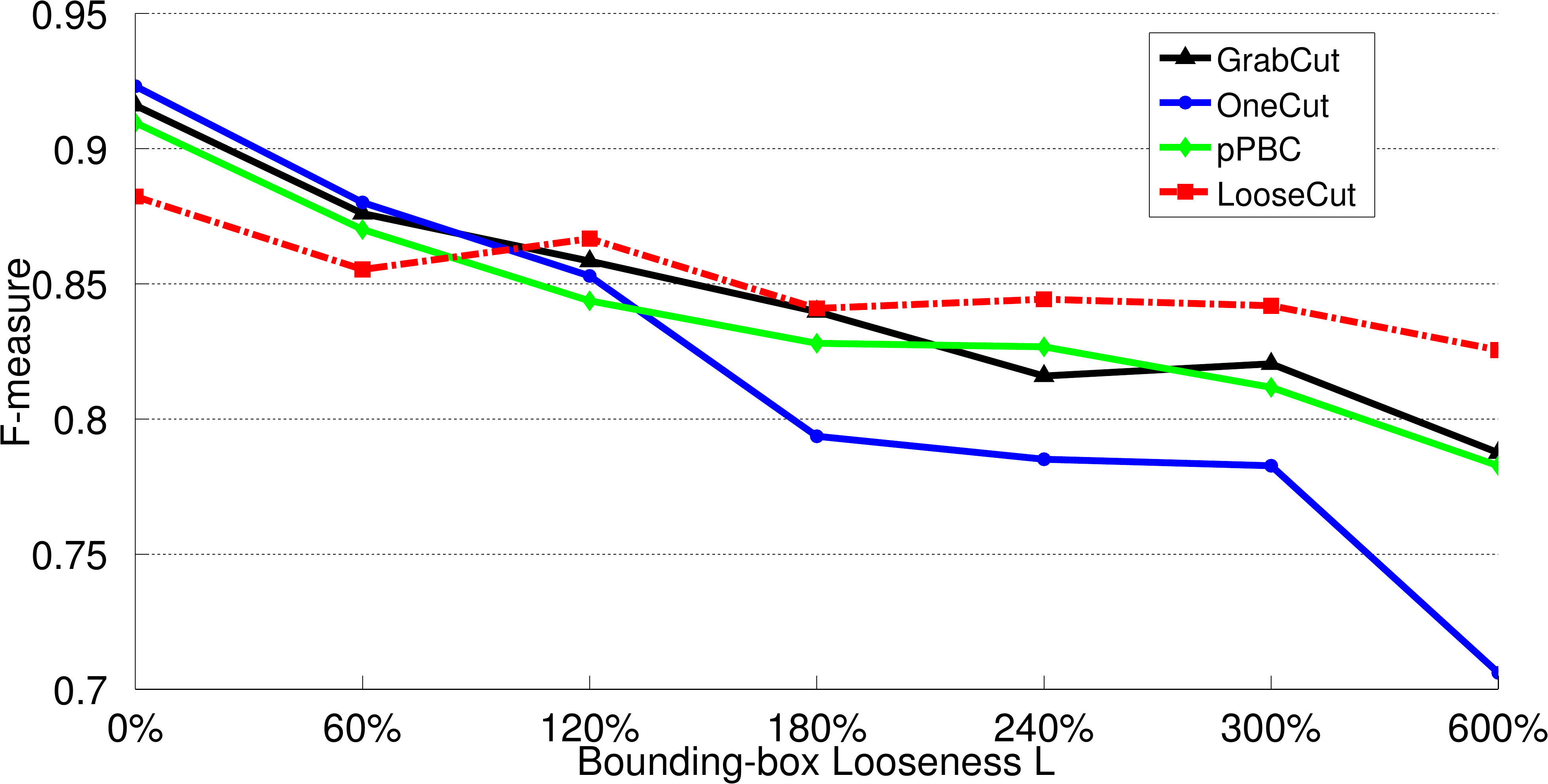} & \includegraphics[width=0.3\textwidth]{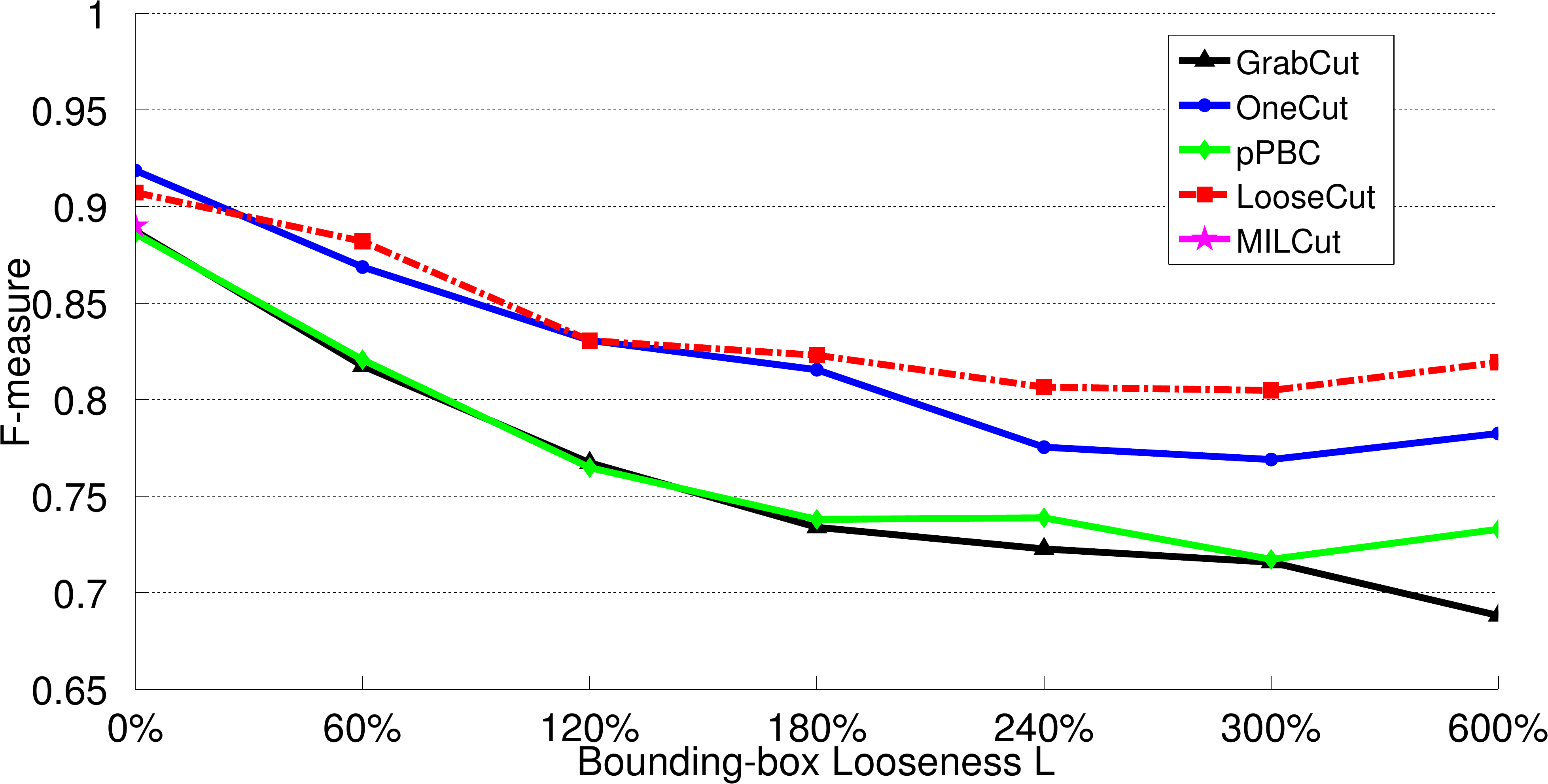} & \includegraphics[width=0.3\textwidth]{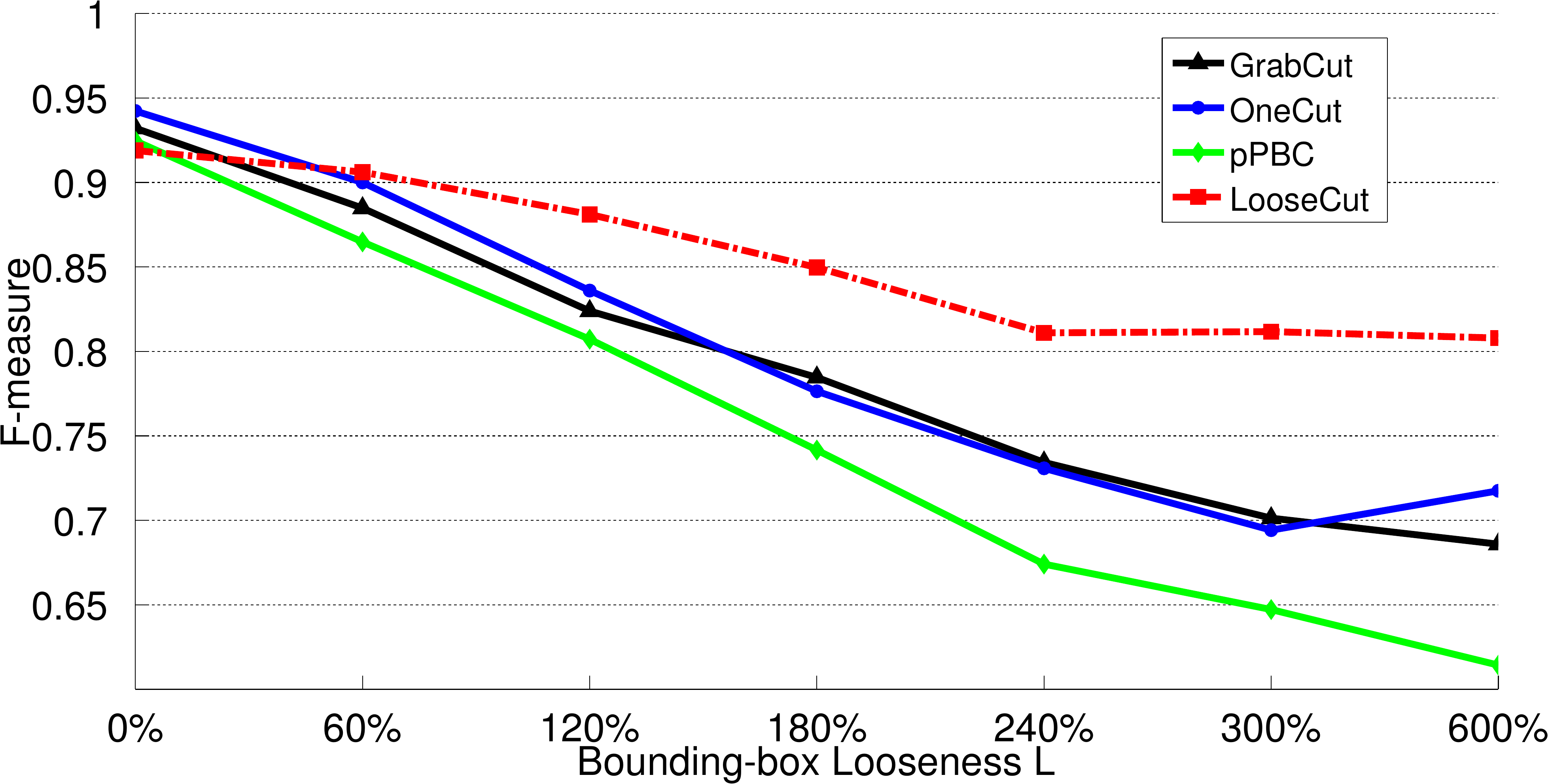}\tabularnewline
\includegraphics[width=0.3\textwidth]{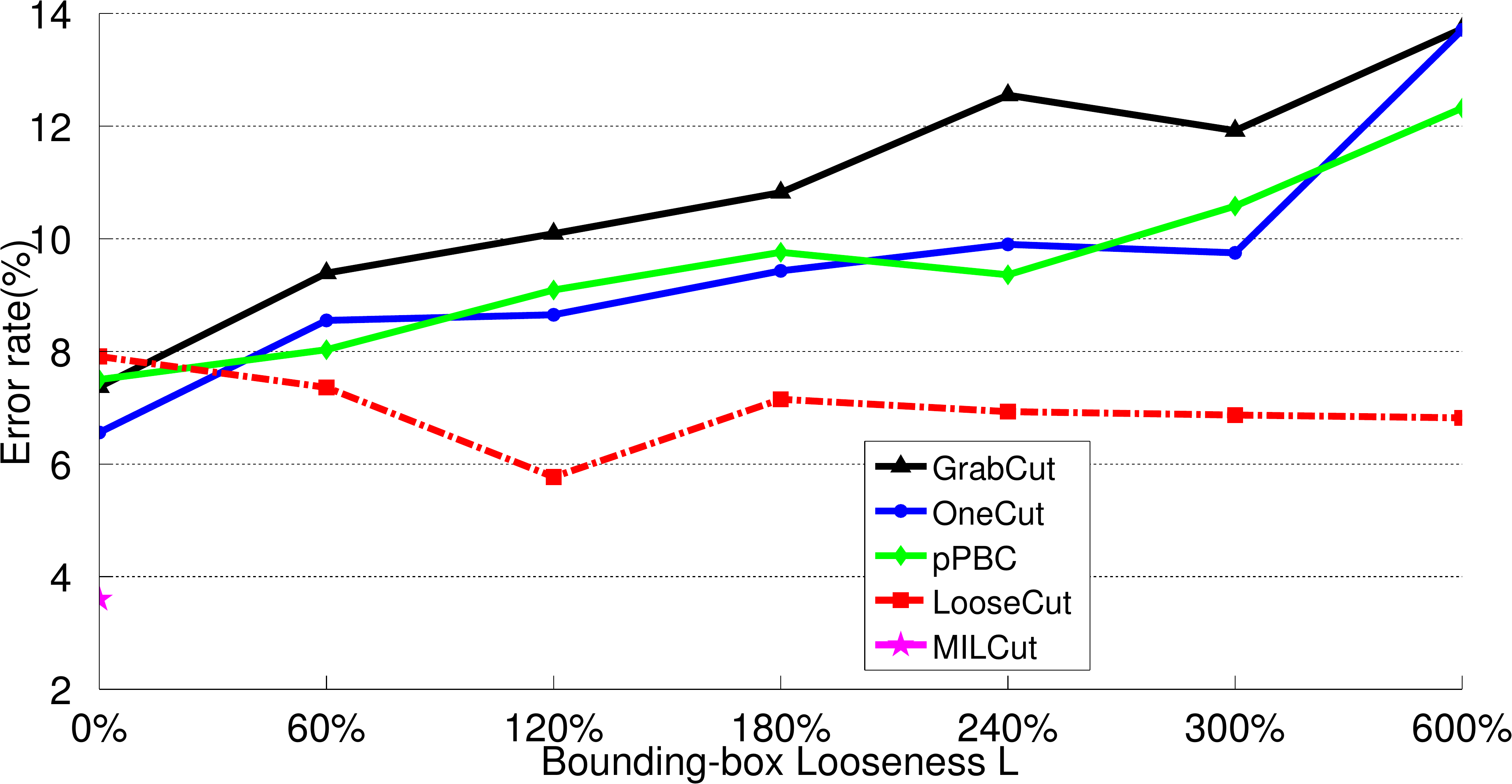} & \includegraphics[width=0.3\textwidth]{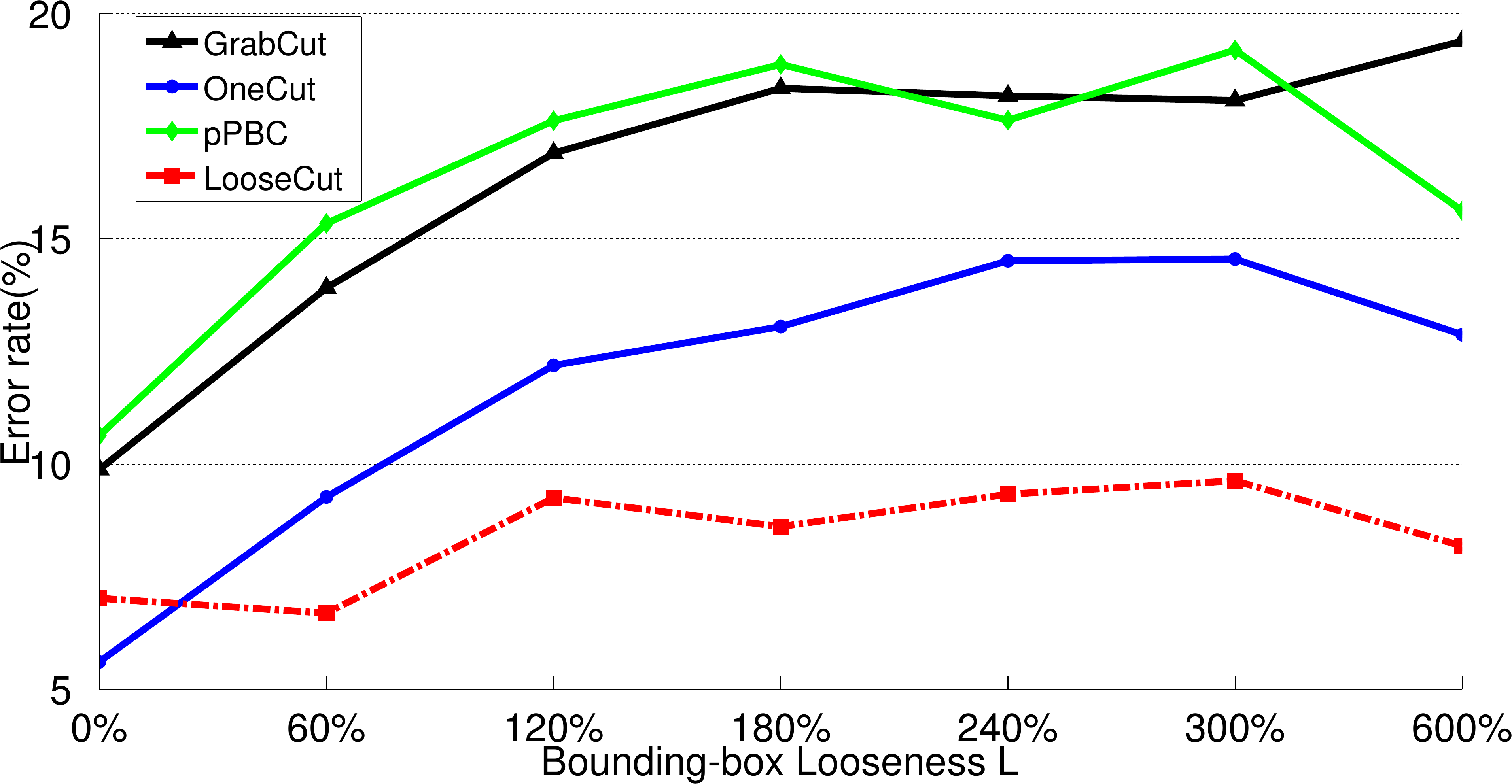} & \includegraphics[width=0.3\textwidth]{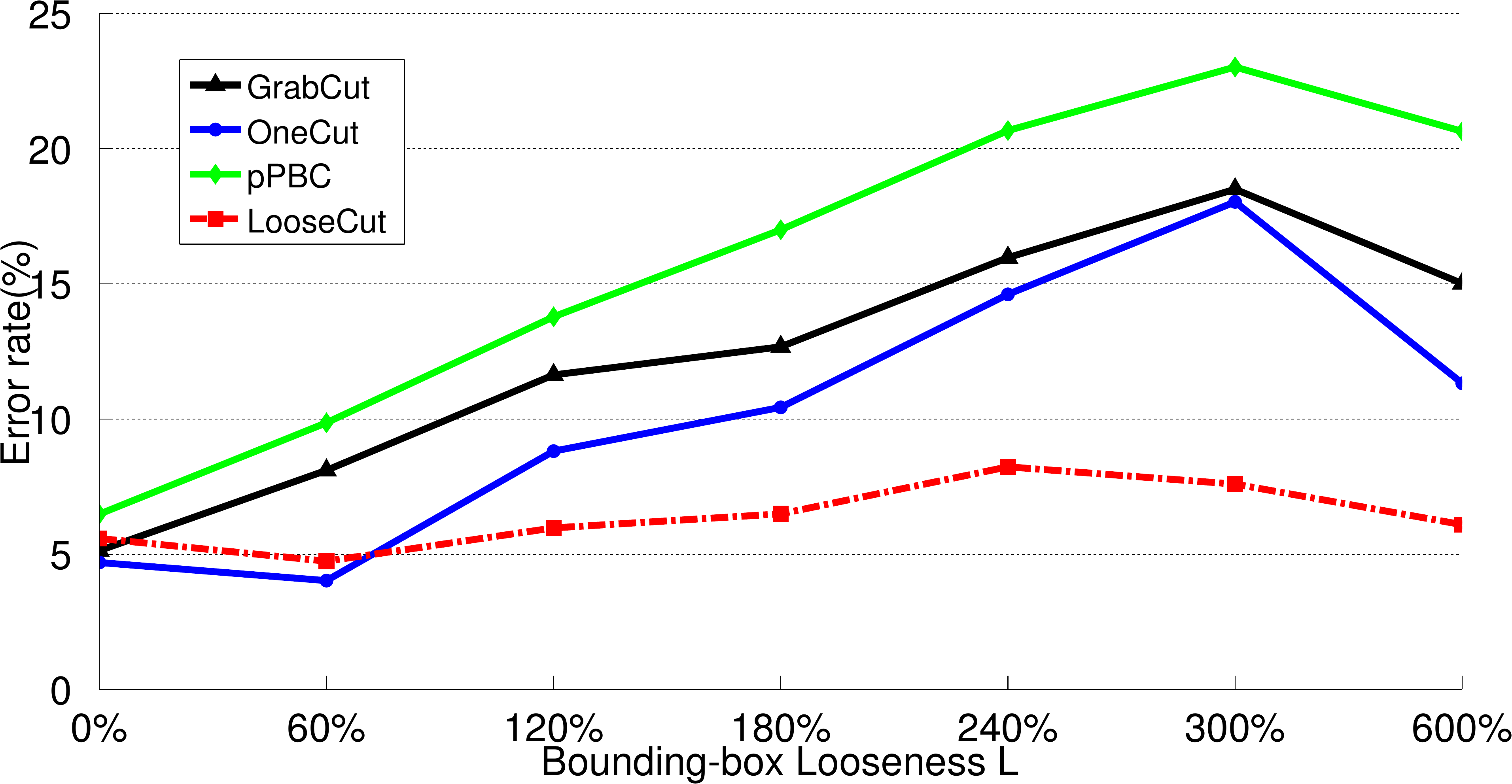}\tabularnewline
GrabCut Dataset & Weizmann Dataset & iCoseg Dataset\tabularnewline
\end{tabular}
\par\end{centering}
\caption{Interactive image segmentation performance (top: F-measure; bottom: Error Rate) on three widely used datasets.}
\label{fig:score}
\end{figure*}

\begin{table*}[htbp]
\begin{centering}
\begin{tabular}{c||c|c|c|c|c|c|c|c}
\hline
\multirow{2}{*}{Methods} & \multicolumn{2}{c|}{$L=0\%$ } & \multicolumn{2}{c|}{$L=120\%$ } & \multicolumn{2}{c|}{$L=240\%$ } & \multicolumn{2}{c}{$L=600\%$ }\tabularnewline
\cline{2-9}
 & F-measure & Error Rate & F-measure & Error Rate & F-measure & Error Rate & F-measure & Error Rate\tabularnewline
\hline
GrabCut  & 0.916 & 7.4 & 0.858  & 10.1 & 0.816  & 12.6 & 0.788  & 13.7\tabularnewline
OneCut  & \textbf{0.923} & 6.6 & 0.853  & 8.7 & 0.785  & 9.9 & 0.706  & 13.7\tabularnewline
pPBC  & 0.910  & 7.5 & 0.844 & 9.1 & 0.827  & 9.4 & 0.783  & 12.3\tabularnewline
MILCut  & -  & \textbf{3.6} & -  & - & -  & - & -  & -\tabularnewline
LooseCut  & 0.882 & 7.9 & \textbf{0.867} & \textbf{5.8} & \textbf{0.844 } & \textbf{6.9} & \textbf{0.826} & \textbf{6.8}\tabularnewline
\hline
\end{tabular}
\par\end{centering}

\centering{}\caption{{\scriptsize{}\label{tabGrabCutDataset}}Segmentation performance
  on GrabCut dataset with bounding boxes of different looseness.}
\end{table*}

\begin{figure*}[htbp]
\begin{centering}
  \includegraphics[width=0.9\textwidth]{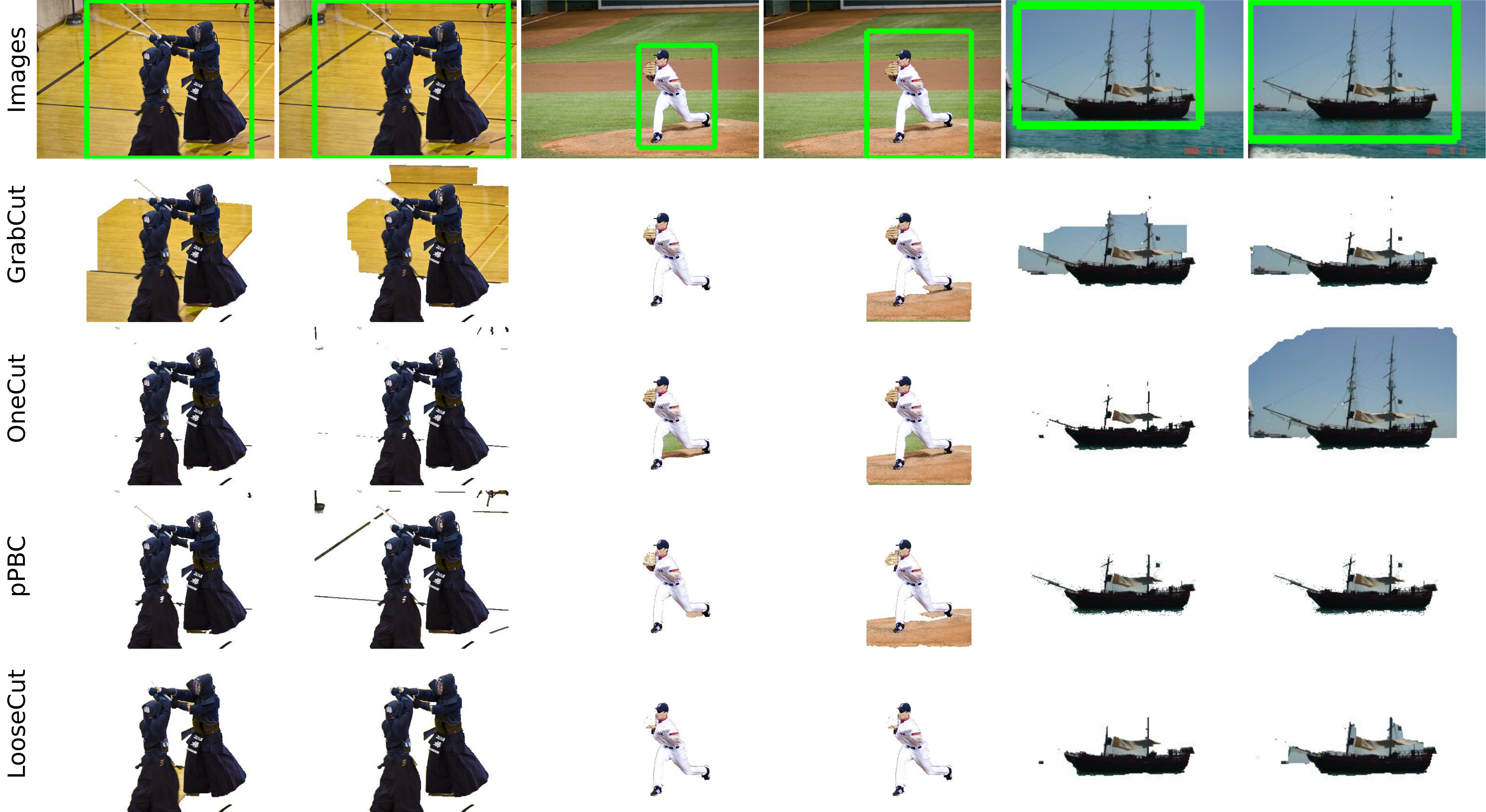}
  \caption{Sample results for interactive image segmentation.}
\vspace{-0.5 em}
\end{centering}
\label{fig4}	
\end{figure*}

\subsection{Interactive Image Segmentation}
In this experiment, we construct bounding boxes with different looseness and
examine the resulting segmentation. As illustrated in Fig.~\ref{figBox}, we
compute the fit box to the ground-truth foreground and slightly dilate it by 10
pixels along four directions, i.e., left, right, up, and down. We take it as the
baseline bounding box with $0\%$ looseness. We then keep dilating this bounding
box uniformly along all four directions to generate a series of looser bounding
boxes -- a box with a looseness $L$ (in percentage) indicates its area increase
by $L$ against the baseline bounding box.  A bounding box will be cropped when
any of its sides reaches the image perimeter. An example is shown in
Fig.~\ref{figBox}.

GrabCut dataset~\cite{rother2004grabcut} consists of 50 images. Nine of them
contain multiple objects while the ground truth is only annotated on a single
object, e.g., ground truth only label one person but there are two people in the loosely bounded box.
Such images are not applicable to test performance change when we enlarge the box looseness.
Therefore, we use the remaining 41 images in our experiments. From Weizmann
dataset~\cite{alpert2007image, borenstein2002class}, we pick a subset of 45
images for testing, by discarding the images where the baseline bounding box has
almost cover the full image and cannot be dilated to construct looser bounding
boxes. For the similar reason, from iCoseg dataset~\cite{batra2010icoseg}, we
select a subset of 45 images for our experiment.

Experimental results on these three datasets are summarized in
Fig.~\ref{fig:score}. In general, the segmentation performance degrades when
the bounding-box looseness increases for both the proposed LooseCut and all the
comparison methods. However, LooseCut shows a slower performance degradation
than the comparison methods. When the looseness is high, e.g., $L=300\%$ or
$L=600\%$, LooseCut shows much higher F-measure and much lower Error Rate than
all the comparison methods. Since MILCut's code is not publicly available, we
only report MILCut's F-measure and Error Rate values with the baseline bounding
boxes on the GrabCut dataset and the Weizmann dataset by copying it from the
original paper. Table~\ref{tabGrabCutDataset} reports the values of F-measure
and Error Rate of segmentation with varying-looseness bounding boxes on GrabCut
dataset. Sample segmentation results, together with the input bounding boxes
with different looseness, are shown in Fig.~\ref{fig4}.

%---------------------------------------------------------------------------------------
\subsection{Unsupervised Video Segmentation}
The goal of unsupervised video segmentation is to automatically segment the
objects of interest from each video frame. The segmented objects can then be
associated across frames to infer the motion and action of these objects. It is
important for video analysis and semantic understanding
\cite{grundmann2010}. One popular approach for unsupervised video segmentation
is to detect a set of object proposals, in the form of bounding boxes
\cite{lee2011key}, from each frame and then extract the objects of interest from
these proposals \cite{zhang2014video}.

In practice, a detected proposal may only cover part of the object of
interest, so we detect a set of object proposals and merge them together to
construct a large mask, which has a better chance to cover the whole object. Clearly, this merged mask may only loosely
bound the object of interest and the object could be extracted by mask based segmentation algorithms.
Specifically, we apply a recent FusionEdgeBox algorithm~\cite{zhou2015feature} to detect top 10 object proposals in
each video frame for the merged mask.

\begin{figure*}[htbp]
\begin{centering}
\includegraphics[width=1\textwidth]{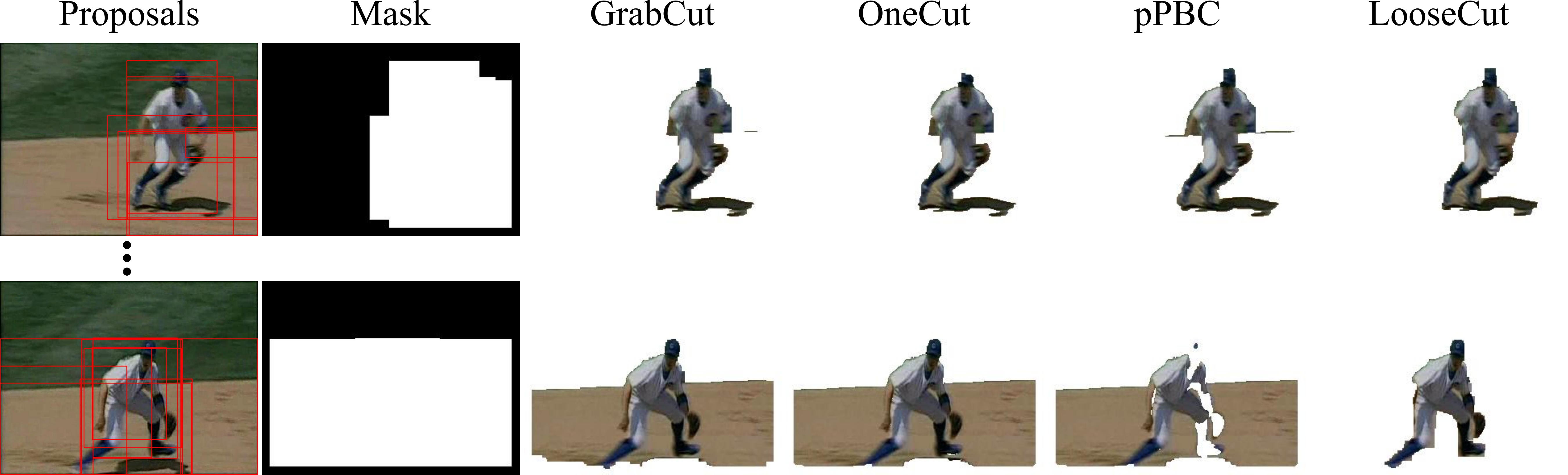}
\caption{\label{fig:samplevideoseg}Sample video segmentation. From left to
  right: top 10 detected object proposals (red rectangles), merged
  mask and different segmentation results.}
\end{centering}
\label{fig5}
\end{figure*}

This experiment is conducted on a subset (21 videos, 657 frames) of JHMDB video dataset~\cite{Jhuang2013towards}. Table~\ref{tab1} shows the unsupervised video segmentation performance, in terms of F-measure and Error Rate averaged over all
the frames. We can see that the proposed LooseCut substantially outperforms GrabCut, OneCut and pPBC in this task. Sample video
segmentation results are shown in Fig.~\ref{fig:samplevideoseg}.

\begin{table}[htbp]
\begin{centering}
{\scriptsize{}}%
\begin{tabular}{c||c|c}
\hline
\scriptsize{}Methods & \scriptsize{}F-measure  & \scriptsize{}Error Rate\tabularnewline
\hline
\scriptsize{}FusionEdgeBox Mask & \scriptsize{}0.35 & \scriptsize{}77.0\tabularnewline
\hline
\scriptsize{}GrabCut  & \scriptsize{}0.55 & \scriptsize{}30.5\tabularnewline
\scriptsize{}OneCut  & \scriptsize{}0.58 & \scriptsize{}25.1\tabularnewline
\scriptsize{}pPBC  & \scriptsize{}0.54 & \scriptsize{}31.6\tabularnewline
\scriptsize{}LooseCut  & \scriptsize{}\textbf{0.64} & \scriptsize{}\textbf{17.0}\tabularnewline
\hline
\end{tabular}
\par\end{centering} {\scriptsize \par}
\centering{}\caption{\label{tab1}Unsupervised video segmentation performance.}
\end{table}

%----------------------------------------------------------------------------------
\subsection{Image Saliency Detection}

\begin{figure*}[htbp]
\begin{center}
  \includegraphics[width=0.7\textwidth]{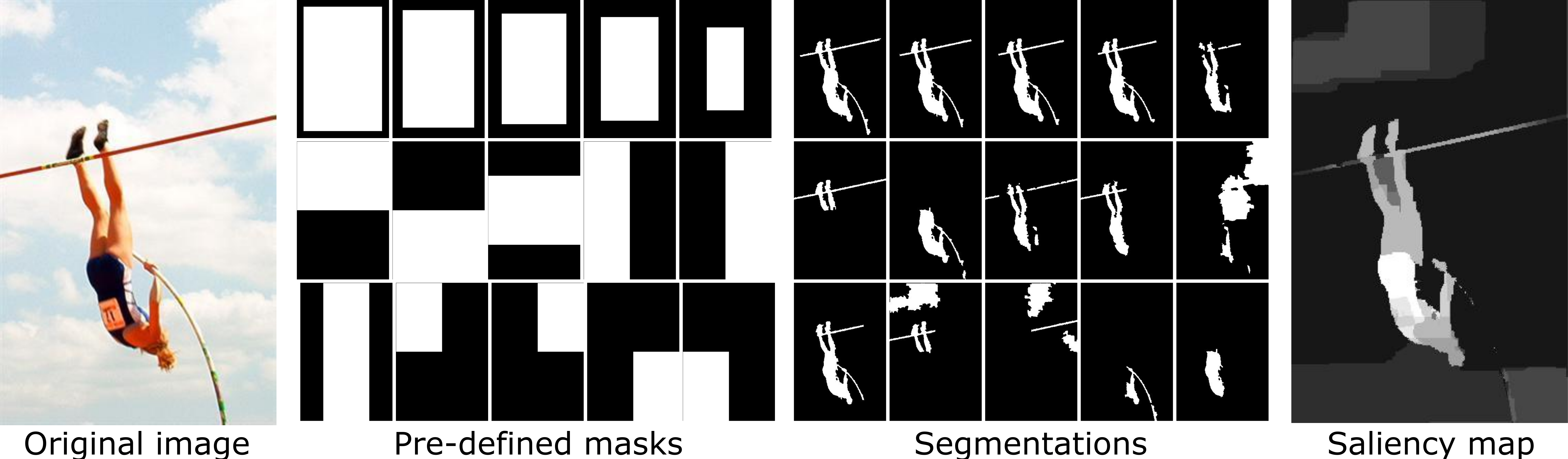}
\end{center}
\vspace{-1 em}
\caption{Segmentation based saliency detection.}
\label{fig6}	
\end{figure*}

Recently, GrabCut has been used to detect the salient area from an image
\cite{li2013co}. As illustrated in Fig.~\ref{fig6}: a set of pre-defined bounding boxes
are overlaid to the input image and with each bounding box, GrabCut is applied
for a foreground segmentation. The probabilistic saliency map is finally
constructed by combining all the foreground segmentations. In this experiment,
it is clear that many pre-defined bounding boxes are not tight.

% to-do: add segmentation results

\begin{figure*}[htbp]
\begin{center}
\includegraphics[width=2.2cm,height=2.8cm]{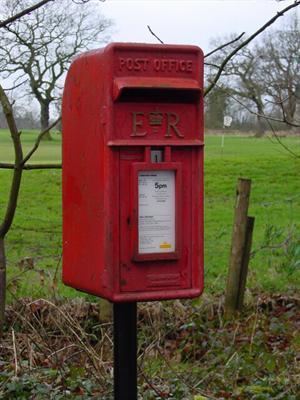}
\includegraphics[width=2.2cm,height=2.8cm]{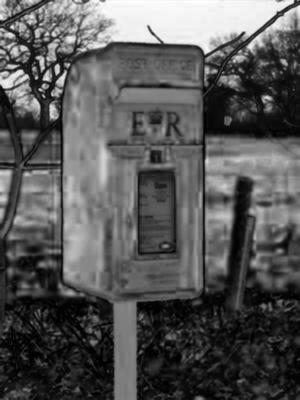}
\includegraphics[width=2.2cm,height=2.8cm]{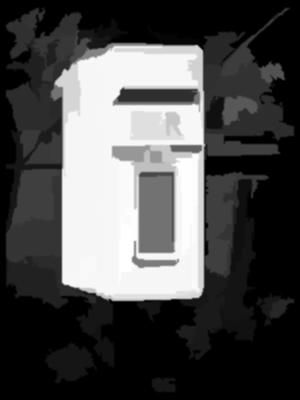}
\includegraphics[width=2.2cm,height=2.8cm]{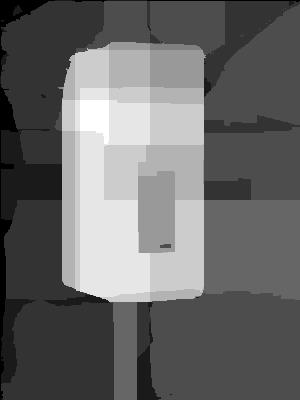}
\includegraphics[width=2.2cm,height=2.8cm]{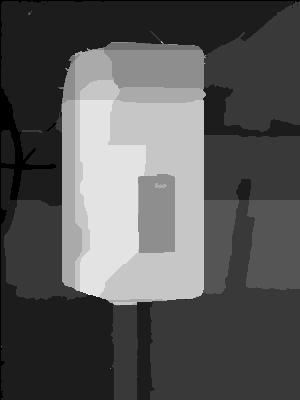}
\includegraphics[width=2.2cm,height=2.8cm]{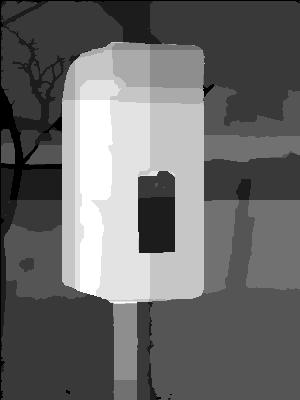}
\includegraphics[width=2.2cm,height=2.8cm]{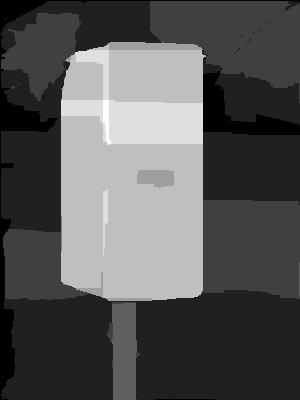}

\includegraphics[width=2.2cm,height=2.8cm]{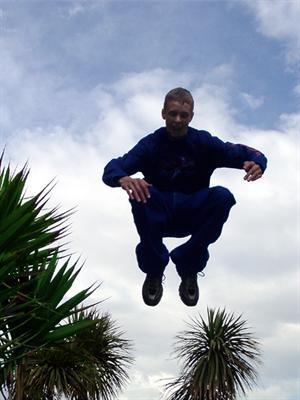}
\includegraphics[width=2.2cm,height=2.8cm]{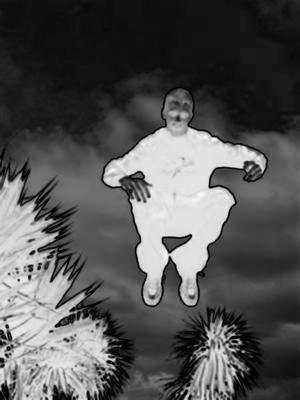}
\includegraphics[width=2.2cm,height=2.8cm]{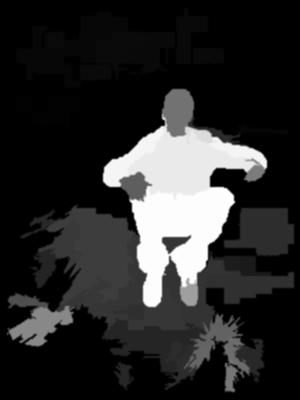}
\includegraphics[width=2.2cm,height=2.8cm]{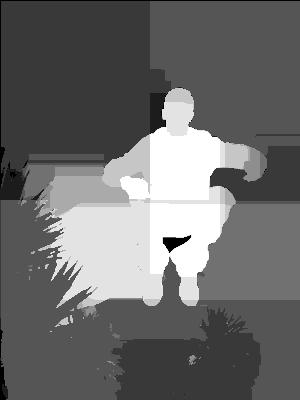}
\includegraphics[width=2.2cm,height=2.8cm]{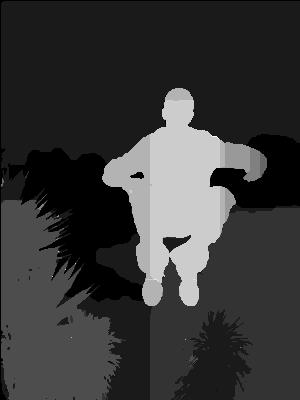}
\includegraphics[width=2.2cm,height=2.8cm]{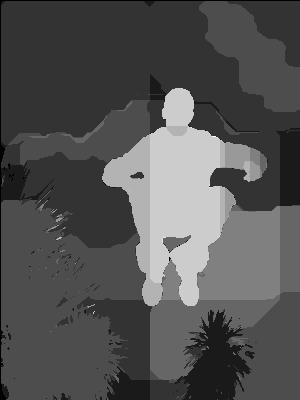}
\includegraphics[width=2.2cm,height=2.8cm]{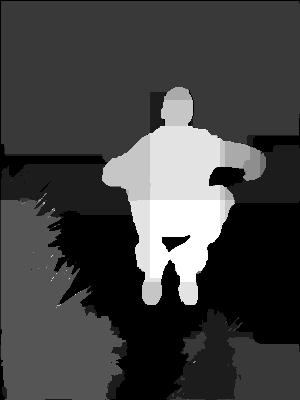}

\includegraphics[width=2.2cm,height=2cm]{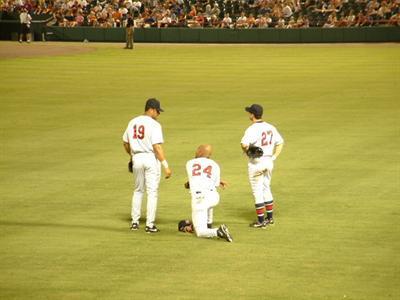}
\includegraphics[width=2.2cm,height=2cm]{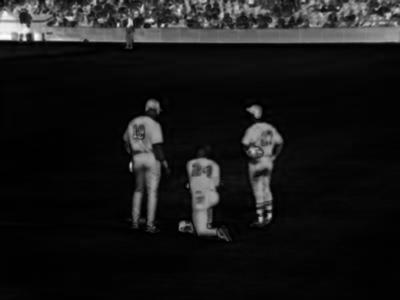}
\includegraphics[width=2.2cm,height=2cm]{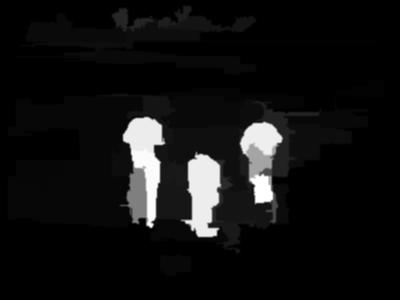}
\includegraphics[width=2.2cm,height=2cm]{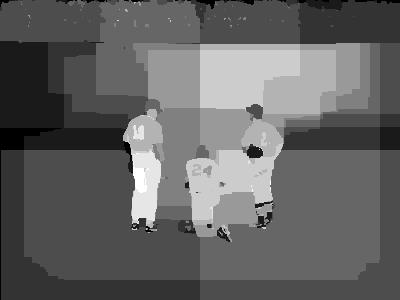}
\includegraphics[width=2.2cm,height=2cm]{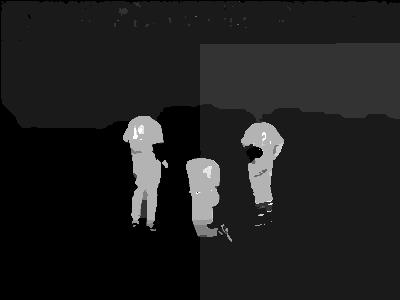}
\includegraphics[width=2.2cm,height=2cm]{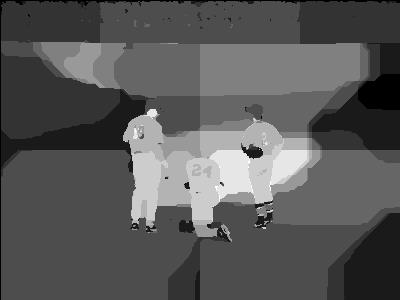}
\includegraphics[width=2.2cm,height=2cm]{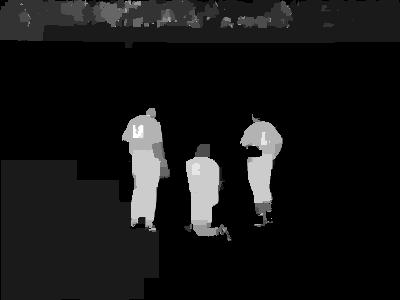}

{\scriptsize \hspace{0.05cm} Image \hspace{1.4cm} FT \cite{achanta2009frequency} \hspace{1.6cm}
RC \cite{Cheng2015PAMI}  \hspace{1.4cm} GrabCut  \hspace{1.4cm}  OneCut \hspace{1.6cm} pPBC \hspace{1.4cm} LooseCut}	
\caption{Sample saliency detection results.}
\label{fig7}	
\end{center}
\end{figure*}

In this experiment, out of 1000 images in the Salient Object
dataset~\cite{achanta2009frequency}, we randomly select 100 images for
testing. 15 pre-defined masks are shown in Fig.~\ref{fig6}.  For quantitative
evaluation, we follow~\cite{achanta2009frequency} to binarize a resulting
saliency map using an adaptive threshold (two times the mean
saliency of the map). Table~\ref{tab2} reports the precision, recall and
F-measure of saliency detection when using GrabCut, OneCut, pPBC, and LooseCut
for foreground segmentation. We also include comparisons of
two state-of-the-art saliency detection methods that do not use pre-defined
masks, namely FT \cite{achanta2009frequency} and RC \cite{Cheng2015PAMI}.
Sample saliency detection results are shown in Fig.~\ref{fig7}.

We can see that LooseCut outperforms GrabCut, OneCut and pPBC in this
task. It also outperforms FT which does not use bounding-box based segmentation.
RC~\cite{Cheng2015PAMI} achieves the best performance for saliency detection,
because it combines more complex saliency cues than segmentation based
approach.

\begin{table}[htbp]
\begin{centering}
{\scriptsize{}}%
\begin{tabular}{c||c|c|c}
\hline
\scriptsize{}Methods & \scriptsize{}Precision  & \scriptsize{}Recall  & \scriptsize{}F-measure \tabularnewline
\hline
\scriptsize{}FT \cite{achanta2009frequency}  & \scriptsize{}0.75 & \scriptsize{}0.57 & \scriptsize{}0.61\tabularnewline
\scriptsize{}RC \cite{Cheng2015PAMI}  & \scriptsize{}0.86 & \scriptsize{}0.85 & \scriptsize{}0.84\tabularnewline
\hline
\scriptsize{}GrabCut  & \scriptsize{}0.85 & \scriptsize{}0.61 & \scriptsize{}0.67\tabularnewline
\scriptsize{}OneCut  & \scriptsize{}\textbf{0.86} & \scriptsize{}0.76 & \scriptsize{}0.77\tabularnewline
\scriptsize{}pPBC  & \scriptsize{}0.84 & \scriptsize{}0.66 & \scriptsize{}0.69\tabularnewline
\scriptsize{}LooseCut  & \scriptsize{}0.84 & \scriptsize{}\textbf{0.78} & \scriptsize{}\textbf{0.78}\tabularnewline
\hline
\end{tabular}
\par\end{centering} {\scriptsize \par}
\centering{}\caption{\label{tab2}Performance of saliency detection.}
\end{table}

\begin{table*}[htbp]
\begin{centering}
\begin{tabular}{c||c|c|c|c|c|c}
\hline
\multirow{2}{*}{Methods}   & \multicolumn{2}{c|}{GrabCut Dataset}                             & \multicolumn{2}{c|}{Weizman Dataset}                             & \multicolumn{2}{c}{iCoseg Dataset}
\\  \cline{2-7}                & F-measure & Error Rate & F-measure & Error Rate & F-measure & Error Rate
\\ \hline
LooseCut w/o proposed constraint \& term                & 0.788      & 13.7          & 0.688          & 19.4           & 0.686         & 15.0      \\
LooseCut w/o global similarity constraint           & 0.801      & 12.0             & 0.709          & 17.9        & 0.691         & 14.8      \\
LooseCut w/o label consistency term           & 0.822      & 7.3           & 0.836          & 7.4            & 0.806         & 6.3       \\
LooseCut     & \textbf{0.826}      & \textbf{6.8}   & \textbf{0.841}         & \textbf{6.6}        & \textbf{0.808}        & \textbf{6.1}        \\  \hline
\end{tabular}
\par\end{centering}
\centering{}\caption{\label{tabEffectsHighOrder} The usefulness of the proposed global similarity constraint and the label consistency term in LooseCut.}
\vspace{-1 em}
\end{table*}

%----------------------------------------------------------------------------------
\subsection{Additional Results \label{sec:additional}}
In this section, we report additional results that justify the usefulness of the
global similarity constraint and the label consistency term, the running time of the proposed algorithm and possible
failure cases.

We run experiments on the three image segmentation datasets when $L=600\%$ by removing the global similarity constraint and/or the label consistency term,
together with their corresponding optimization steps in the proposed LooseCut algorithm. The
quantitative performance is shown in Table~\ref{tabEffectsHighOrder}. 
We can see that both the global similarity constraint and the label consistency term help
improve the segmentation performance.
The global similarity constraint helps improve the segmentation performance more significantly than the label consistency term.

For the running time, we test LooseCut and all the comparison methods on a PC
with Intel 3.3GHz CPU and 4GB RAM. We compares their running time for different
image size. In this experiment, OneCut only has one iteration, and the
iterations of GrabCut and LooseCut are stopped until convergence or a maximum 10
iterations is reached.  As shown in Table~\ref{tabTime}, if the image size is
less than $512\times 512$, the running time of three algorithms are very
close. For large images, LooseCut and OneCut takes more time than GrabCut.  In
general, LooseCut still shows reasonable running time. Our current LooseCut code
is implemented in Matlab and C++, and it can be substantially optimized for
speed.

\begin{table}[htbp]
\begin{centering}
{\scriptsize{}}%
\begin{tabular}{c||c|c|c|c|c}
\hline
{\scriptsize{}Methods} & {\scriptsize{}64{*}64 } & {\scriptsize{}128{*}128 } & {\scriptsize{}256{*}256 } & {\scriptsize{}512{*}512 } & {\scriptsize{}1024{*}1024 }\tabularnewline
\hline
{\scriptsize{}GrabCut } & {\scriptsize{}0.16 } & {\scriptsize{}0.28 } & {\scriptsize{}1.47 } & {\scriptsize{}3.81 } & {\scriptsize{}25.21 }\tabularnewline
{\scriptsize{}OneCut } & {\scriptsize{}0.03 } & {\scriptsize{}0.09 } & {\scriptsize{}0.49 } & {\scriptsize{}5.72 } & {\scriptsize{}77.80 }\tabularnewline
{\scriptsize{}pPBC } & {\scriptsize{}0.14 } & {\scriptsize{}0.37 } & {\scriptsize{}2.70 } & {\scriptsize{}26.14 } & {\scriptsize{}305.60 }\tabularnewline
{\scriptsize{}LooseCut } & {\scriptsize{}0.32 } & {\scriptsize{}0.43 } & {\scriptsize{}1.68 } & {\scriptsize{}7.63 } & {\scriptsize{}66.52 }\tabularnewline
\hline
\end{tabular}
\par\end{centering}{\scriptsize \par}
\centering{}\caption{\label{tabTime}Running time (in seconds) with increasing image size.}
\vspace{-1 em}
\end{table}

\begin{figure}[htbp]
\begin{center}
\includegraphics[width=1\columnwidth]{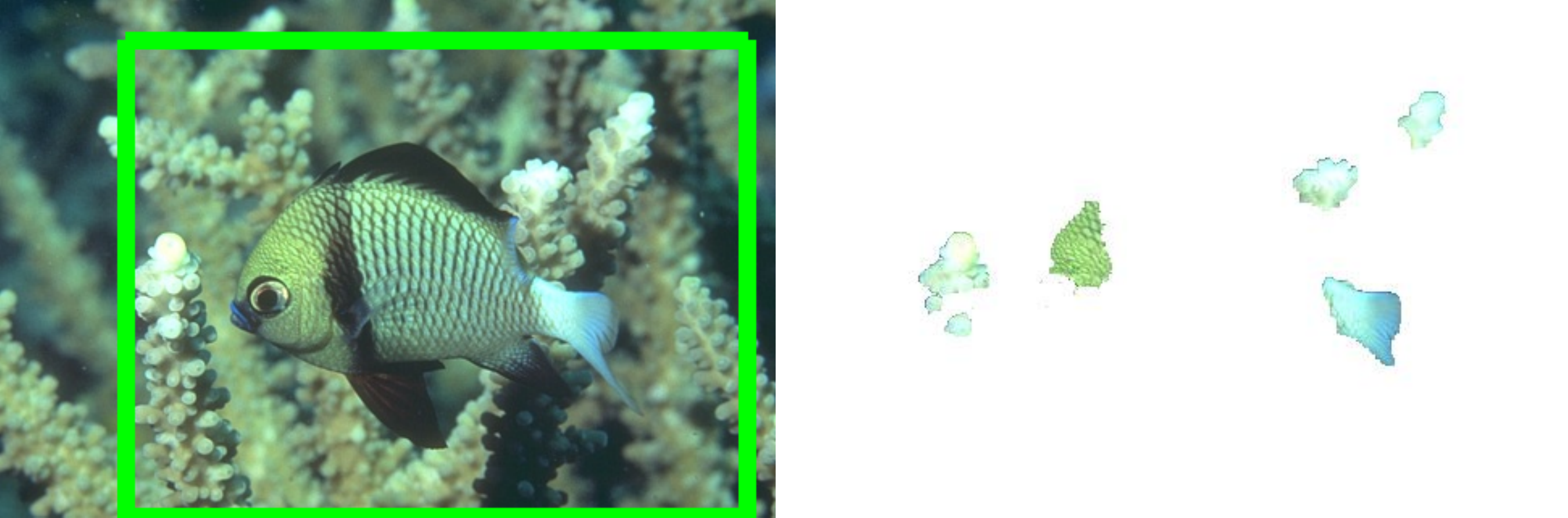}
\caption{Failure cases of LooseCut.}
\vspace{-2 em}
\label{fig9}	
\end{center}
\end{figure}

%-------------------------------------------------------------------------------------
Due to the proposed global similarity constraint and label consistency term, LooseCut may fail when the foreground and
background show highly similar appearances, as shown in Fig.~\ref{fig9}.
%where the foreground object shows similar colors and texture to the
%background.

%-------------------------------------------------------------------------
\section{Conclusion\label{sec:end}}
This paper proposed a new LooseCut algorithm for interactive image segmentation
by taking a loosely bounded box. We further introduced a global similarity constraint
and a label consistency term into MRF model. We developed an
iterative algorithm to solve the new MRF model. Experiments on three
image segmentation datasets showed the effectiveness of LooseCut against several
state-of-the-art algorithms. We also showed that LooseCut can be used to
enhance the important applications of unsupervised video segmentation and image
saliency detection.
%-------------------------------------------------------------------------
{\small
\bibliographystyle{ieee}
\bibliography{Hongkai_bib}
}

\end{document}